\documentclass[11pt]{article}
\usepackage[margin=1.03in]{geometry}
\usepackage{amsmath,amssymb,amsthm}
\usepackage[hidelinks]{hyperref}
\newcommand{\VC}{\operatorname{VC}}
\newcommand{\LD}{\operatorname{Ldim}}
\newcommand{\FF}{\mathbb F}
\newcommand{\HH}{\mathcal H}
\newcommand{\BB}{\mathcal B}
\newtheorem{theorem}{Theorem}[section]
\newtheorem{lemma}[theorem]{Lemma}
\newtheorem{definition}[theorem]{Definition}
\newtheorem{feature}[theorem]{Feature}
\title{Refuting the Log-Star Conjecture for Private Learning}
\author{Uri Stemmer\thanks{Tel Aviv University. \texttt{u@uri.co.il}} \and Eliad Tsfadia\thanks{Bar-Ilan University. \texttt{eliad.tsfadia@biu.ac.il}}}
\date{}
\begin{document}
\maketitle
\begin{abstract}
Several researchers have conjectured that the sample complexity of PAC learning a class $\HH$ under differential privacy is at most ${\rm poly}(\VC(\HH),\, \log^*(\LD(\HH)))$, where $\LD$ is the Littlestone dimension. We refute this conjecture by showing that the private sample complexity can grow superlinearly in the Littlestone dimension, even when the VC dimension is constant. This rules out even a generic upper bound of the form ${\rm poly}(\VC(\HH),\, \log(\LD(\HH)))$. Using the same technique, we also show that the private sample complexity of a class $\HH$ of VC dimension two can be as large as $\Omega(\log\log\log|\HH|)$.
\end{abstract}

\section{Introduction}
How many more examples does a PAC learner need when it must protect the privacy of its training data? Without privacy constraints, it is well known that the sample complexity of PAC learning a binary concept class $\HH$ is of order $\VC(\HH)$ (at constant accuracy and confidence). Kasiviswanathan et al.\ introduced private PAC learning and showed that $O(\log|\HH|)$ examples suffice for every finite class, even under \emph{pure} differential privacy~\cite{KLNRS}. Under \emph{approximate} differential privacy, Alon et al.~\cite{ABlMM} showed that finite Littlestone dimension characterizes learnability and obtained a sample lower bound of $\Omega(\log^* \LD(\HH))$. The general upper bound went from doubly exponential in $\LD(\HH)$ to polynomial~\cite{ABlMM,GGKM,Lyu} and, most recently, to nearly linear in the work of Moran and Yan~\cite{MY}.

When $\VC(\HH)$ is small, how much of this dependence on $\LD(\HH)$ is necessary? Several researchers have considered the conjecture that $\operatorname{poly}(\VC(\HH),\log^* \LD(\HH))$ samples suffice. In particular: (1) Yan~\cite{Yan} conjectured such a log-star dependence on the Littlestone dimension and established it for classes with VC dimension one; (2) Nissim et al.~\cite{NST} explicitly asked whether $\operatorname{poly}(\VC(\HH),\log^* \LD(\HH))$ samples suffice in general; and (3) Dharangutte and Sanyal~\cite{DharS} explicitly identified this bound as a possibility left open by their lower bounds. We show the following counterexample.

\begin{theorem}\label{main}
There is a sequence of finite binary concept classes $\HH=\{\HH_L\}_{L\geq4}$ with $\VC(\HH_L)=2$ and $\LD(\HH_L)=L$ such that the following holds. Every realizable $(\frac{1}{10},\frac{1}{10})$-PAC learner for $\HH_L$ that uses $n$ samples and satisfies $(1,n^{-1.1})$-DP must have $n=\Omega(L\log L)$, even if the learner is improper.\footnote{Under $(1,n^{-1})$-DP, our proof gives $n=\Omega(L)$.}
\end{theorem}

A corollary of our result is that the general upper bound of Moran and Yan~\cite{MY}, which is $O(L\log L)$, is tight up to constant factors even among classes of VC dimension exactly two.

We also use our technique to obtain a lower bound in terms of the cardinality of the class.

\begin{theorem}\label{thm:card}
There is a sequence of finite binary concept classes $\{\HH'_m\}$, defined for all sufficiently large $m$, with $\VC(\HH'_m)=2$ and $\log\log\log|\HH'_m|=\Theta(m\log m)$ such that the following holds. Every realizable $(\frac{1}{10},\frac{1}{10})$-PAC learner for $\HH'_m$ that uses $n$ samples and satisfies $(1,n^{-1.1})$-DP must have $n=\Omega(m\log m)=\Omega(\log\log\log|\HH'_m|)$, even if the learner is improper.\footnote{Under $(1,n^{-1})$-DP, our proof gives $n=\Omega(m)$.}
\end{theorem}

Theorem~\ref{thm:card} partially resolves Open Question 2 of Nissim et al.~\cite{NST}, concerning the dependence of the private sample complexity on the size of the class.

\paragraph{Comparison with prior work.}
Dharangutte and Sanyal~\cite{DharS} exhibited a polynomial separation between the private sample complexities of classes with identical VC and Littlestone dimensions, at the same privacy and utility parameters. Their result rules out a characterization in terms of these two dimensions alone up to polylogarithmic factors, while leaving open a characterization up to polynomial factors. They explicitly left open whether $\operatorname{poly}(\VC(\HH),\log^*\LD(\HH))$ samples suffice for every class $\HH$. Theorem~\ref{main} rules out even the more generous upper bound $\operatorname{poly}(\VC(\HH),\log\LD(\HH))$.

\subsection{Proof idea}\label{sec:idea}
We gradually introduce the ideas behind Theorem~\ref{main}. The description in this section is informal and omits some details of the actual construction.

\subsubsection{Warmup}

Consider a target function consisting of a straight line $\ell$ in the plane. An input is a point $p$ in the plane, and its label is $1$ exactly when $p$ lies on $\ell$. Two distinct points on a line determine it, so the class of lines in the plane has VC dimension 2. Of course, this class is too easy for our purposes: its Littlestone dimension is also 2, and it can be privately learned using a constant sample complexity.\footnote{For example, via the sample-and-aggregate technique, paired with stability-based arguments.}

\paragraph{Add a second sheet of paper.}
Draw a line $\ell_1$ on the first sheet. On the second sheet, draw the same line shifted vertically by some amount $d$. The shift is part of the target function. An input on the first sheet is simply a point. An input on the second sheet gives both a point and a claimed value for $d$. Its label is $1$ exactly when the point lies on the second line and the claimed shift is correct.

Suppose we see the positive point $(1,4)$ on the first sheet. Then we see the positive point $(3,11)$ on the second sheet, with claimed shift $d=3$. Because the second example is positive, we know the shift really is $3$. Moving its point down by $3$, we learn that $(3,8)$ lies on the first line. We now have two points on the first line, so we know both lines.

There is an asymmetry here. Even if we had already learned the first line perfectly, it would tell us nothing about how far the second line had been shifted. A positive example on the second sheet, however, tells us the shift and lets us move its point back to the first sheet.

\paragraph{Now imagine a sequence of sheets.}
Each new line is obtained by shifting the preceding line. An input on sheet $j$ carries a list of the first $j-1$ shifts. As before, a positive label certifies both the point and its list. We give different sheets disjoint sets of \emph{horizontal} coordinates, so that points from different sheets always give observations at different coordinates. Moving a point up or down preserves its horizontal coordinate, so two such points remain distinct when we move them onto the same sheet.

\paragraph{Why can the VC dimension stay at two even with many sheets?} Take any three distinct examples that can all be positive. Pick the example on the highest-numbered sheet and one of the other two. The first supplies all the shifts needed to put both points on the same sheet. Those two points determine its line, and the shifts determine all the earlier lines. In particular, they settle the label of the third example.

Yet even if we know the lines on sheets $1,\ldots,j$ exactly, we still do not know the line on sheet $j+1$. Its shift can be chosen freely. Once we see a positive example on sheet $j+1$, its list tells us the shift and lets us translate the point back to any of the first $j$ sheets.

This construction can be formalized to give a class with VC dimension two and arbitrarily large Littlestone dimension. Yet it still does not give the privacy lower bound we seek. To see why, suppose every example carries the correct list of shifts. The list on each example tells us exactly how to move its point back to the first sheet. Once we know the line on that sheet, we can predict on every sheet, however long the sequence is.

\subsubsection{Hiding the sequence of sheets}
To make the problem hard under differential privacy while keeping it easy to learn without privacy, we hide which sequence of sheets supplies the examples. Imagine a public collection of $N\gg m$ sheets from which the training examples can be drawn. Each of the $N$ sheets has a different name printed at the top, and that name is a nonempty proper subset $B\subsetneq U$, where $U$ is a finite universe of words. There is also a sheet named $U$ itself, which serves as a common reference sheet. For example, one sheet might be named by the singleton $\{u\}$, and another by a set $B$ containing $u$ and several other words.

The sheet names are chosen so that every maximal nested sequence among these $N$ names contains exactly $m=L-2$ sets. A given sheet can belong to many such sequences, but it occupies the same position in every sequence to which it belongs. We secretly choose one of these sequences uniformly and write its sheet names in increasing order,
\[
F_1\subsetneq F_2\subsetneq\cdots\subsetneq F_m.
\]
The learner sees the name on each sheet it encounters. That name tells it the sheet's position in the sequence, but generally leaves many possibilities for the other sheets. For example, if the second sheet is named $B$ and contains the words $u$ and $v$, the same sheet $B$ could have been preceded by the sheet named $\{u\}$ or by the sheet named $\{v\}$.

As before, the target concept specifies a line and a list for each sheet. In the warm-up, each sheet had a unique sequence leading to it, so its list could simply record the shifts along that sequence. Now several sequences can lead to the same sheet $B$, so the lists are built differently, as we now describe. The target concept gives each of the $N$ sheets a secret \emph{tag} of its own. These tags play two roles:
\begin{enumerate}
    \item First, they \emph{collectively} determine all the lines in the target concept. The line on a sheet $B$ is the line on the reference sheet $U$ moved by a shift, and computing this shift involves the tags of all the sheets that contain $B$ (including $B$'s own tag).
    \item Second, they determine the lists: the list on a sheet $B$ contains every tag \emph{except} those of the sheets that contain $B$.
\end{enumerate}

In particular, $B$'s list does not contain $B$'s tag. More generally, it hides precisely the tags from the first item, so on its own it does not reveal how $B$'s line is shifted from the reference sheet. Nevertheless, computing the \emph{relative} shift from a sheet $A\subseteq B$ to $B$ remains possible: the shifts for $A$ and for $B$ differ only through the sheets that contain $A$ but not $B$, so their difference can be computed from the tags of the sheets that do not contain $B$, that is, from $B$'s list. The list is not enough to move a point to a sheet that is not inside $B$.

This construction helps avoid the issue we ran into before: In the warm-up, we could move every point back to the \emph{same} publicly known first sheet. Here, an example from a sheet $B$ containing $u$ and $v$ lets us move its point to either $\{u\}$ or $\{v\}$, but it does not tell us which one is the first sheet shared by the whole sample.

\paragraph{How can we keep the VC dimension at 2?} We do this by enforcing a special structure on the names of our sheets and their lists. Specifically, we enforce the following two properties:
\begin{enumerate}
    \item Every pair of sheet names has a smallest containing set in the family. Among any three names, some two already determine the smallest set in the family containing all three. That is, for every $B_1, B_2, B_3$ there are $i\neq j$ such that the smallest set containing both $B_i, B_j$ is the smallest set containing $B_1, B_2, B_3$.

    \item The lists themselves are also nested whenever their sheet names are nested: if $B_1\subseteq B_2$, every entry on $B_1$'s list also appears on $B_2$'s list. More generally, if $C$ is the smallest set in the family containing both $B_1$ and $B_2$, then $C$'s list is the union of their lists.
\end{enumerate}

To see that three examples cannot be shattered, take three distinct examples that can all be positive. Pick the two whose sheet names determine the smallest set $C$ containing all three. Any target that labels those two examples positive certifies both their lists. Together the lists give us $C$'s list, which lets us move the two points to sheet $C$. The resulting points are distinct, so they determine the line on $C$. Since the third sheet is contained in $C$, we can now recover its correct list and its line. Thus every target that labels the chosen pair positive gives the third example the same label. Since all three can be positive, that label must be positive. This proves that the VC dimension is at most two.

\paragraph{How do we draw a labeled example?}
We choose the target concept\footnote{It suffices to specify the line and the list on the reference sheet $U$. For every sheet $B$, this list tells us how to shift the line back from $U$ to $B$, and the name $B$ tells us which entries to keep for $B$'s own list.} and the secret sequence\footnote{We choose uniformly among all maximal nested sequences of sheet names.} $F_1\subsetneq\cdots\subsetneq F_m$ independently, once for the entire sample. To draw an example from one of the sheets, choose $j$ uniformly from $1,\ldots,m$ and go to the sheet named $B=F_j$. To choose a point $(x,y)$ on this sheet, we do as follows. First, choose a horizontal coordinate $x$ uniformly from the horizontal coordinates assigned to that sheet. Let $z$ be the height of the target line at $x$ on this sheet, and suppose that it is an integer (in the actual construction, heights are finite-field elements). If $z$ is odd, we set $y=z$ and set the label to be $1$. Otherwise, if $z$ is even, we set $y=z+1$ and set the label to be $0$.

The learner receives the sheet name $B$, the list for sheet $B$, the displayed point $(x,y)$, and its label. Thus the displayed point tells the learner that the line has height either $y-1$ or $y$ at $x$ on sheet $B$, and the label tells it which of the two is correct.

\subsubsection{Where privacy enters}

Without privacy, a single sheet example lets us name a sheet on the hidden sequence. Since the sheets are nested, it also lets us solve an easier task: name any word belonging to the last sheet $F_m$. Indeed, any word on the sheet we see belongs to $F_m$. We first show that this word-finding task is hard under differential privacy. We then show that an accurate binary predictor for our designed class lets us solve this word-finding task (this is the role of the lines and lists), thus showing that the learning task we defined is hard under DP.

\paragraph{The hardness of the word-finding task.}
A crucial property of our construction is that, for every maximal sequence $F_1\subsetneq F_2\subsetneq\cdots\subsetneq F_m$ of proper sheet names, if we hide one $F_j$ and reveal every other sheet in the sequence, then exactly $r=(m+1)^4$ sheets could fill the gap.

\begin{definition}
Put $F_0=\varnothing$. We say that a word in $F_m$ \emph{first appears} at level $j$ if it belongs to $F_j\setminus F_{j-1}$.
\end{definition}

Fix a level $j$. In the training sample, replace every example whose sheet name is $F_j$ by a fixed blank record carrying no sheet name, keeping all other examples. Now suppose we are told every sheet in the hidden sequence except $F_j$. How likely are we to name a word that first appears at level $j$, that is, a word in $F_j\setminus F_{j-1}$? There are still $r$ equally likely choices for the missing sheet. Moreover, different choices add disjoint sets of new words to $F_{j-1}$. Thus, whatever word the algorithm names after the erasure, at most one of the $r$ choices makes it a word that first appears at level $j$. Its chance of doing so is at most $1/r$.

Differential privacy limits how much restoring the examples from level $j$ can increase this probability. The calculation is similar to the analysis of privacy amplification by sampling with replacement. We condition on the number of times level $j$ was sampled, apply group privacy for that many replacements, and average over the binomial distribution of this count. A union bound over the $m$ levels then gives the $\Omega(m\log m)$ lower bound for the privacy parameters in our theorem.

\paragraph{Lifting the hardness to binary predictors.}
Suppose we are given sheet names from the hidden sequence and want to find a word in $F_m$. We choose the target concept ourselves, independently of the sequence, and use it to turn each sheet name into a labeled example. We then run the learner and obtain its predictor $h$. From this point on, we use only $h$ and the target we chose, with no further access to the hidden sequence. Informally, we test the behavior of $h$ on many possible inputs, in a way that lets us statistically rule out candidate sheets for $F_m$. We will not be able to single out $F_m$ itself. But we will rule out every sheet that is disjoint from $F_m$ while keeping $F_m$, and this turns out to be enough to find a word in it.

The test we run on $h$ relies on how the tags shape the lines, so we first make this more precise. The tag of a sheet $D$ has two parts: a \emph{key}, chosen uniformly from a public set of $100(N+2)$ values, and an \emph{adjustment}, which is itself a uniformly random line. To obtain the line on a sheet $B$, start from the line on the reference sheet $U$. Then, for every sheet $D$ other than $U$ that contains $B$, including $B$ itself, add $D$'s adjustment multiplied by the sum of the keys of all the sheets disjoint from $D$.\footnote{Lines are added height by height, so the result is again a line. These are the ``shifts'' described earlier. So a shift may now change the slope of a line as well as its height.} The line on $U$ is also uniformly random, and it appears on no list (similarly, the adjustment of a sheet $D$ is omitted from the lists of the sheets inside $D$). The important feature here is the following:
\begin{feature}[Limited effect of keys]\label{feature}
The key of $D$ affects the line on a sheet $B$ only if $B$ is disjoint from $D$. In particular, it never affects the line on a sheet inside $D$.
\end{feature}
Fix one of the $N$ sheet names $D$ as a candidate for $F_m$. We build a test table with one column for each of the $\approx 100N$ possible keys. To form a column, we change $D$'s key to that value, keep the rest of the target fixed, and recompute the lines and lists. This changes the list on every sheet that shows $D$'s tag, and it can change the lines on sheets disjoint from $D$. We know which column corresponds to the original target.

The table has one row for each sheet $B$, including the reference sheet $U$, for a total of $N+1$ rows. Each cell of the table holds a score: the cell in row $B$ and a given column measures how well $h$ predicts on sheet $B$ when the target is that column's target. To compute it, we go through the horizontal coordinates assigned to $B$. At each coordinate we use that column's target to generate a labeled example. We give $h$ the sheet name, list, and displayed point, and compare its prediction with the correct label. The score is the fraction of coordinates on which $h$ is correct. In each row we select the column with the highest score. If any row selects the original column, we discard the candidate $D$.

To see why discarding $D$ makes sense in this context, suppose first that $D$ is disjoint from $F_m$; we show that $D$ is then likely to be discarded. We assume that $h$ is accurate w.r.t.\ the target concept when sampling from the target distribution. By an averaging argument, $h$ is accurate on at least one of the $m$ sheets in the hidden sequence. Call it $B$. So in the table, the original column has a high score in row $B$. Now consider an alternative column. Since $D$ is disjoint from $F_m$, its key is one of the keys that multiply $F_m$'s adjustment. Changing it therefore changes the line on $B$ by an amount that includes a nonzero multiple of $F_m$'s adjustment. But $h$ knows nothing about $F_m$'s adjustment. It is on the list of no hidden sheet, since $F_m$ contains all of them. The lines on the hidden sheets do depend on it, but always in one fixed combination with the line on $U$, and that combination reveals nothing about the adjustment alone. Consequently, in each alternative column the line on $B$ has a uniformly random slope and intercept, even given $h$ and the list it sees on $B$. (This is the main reason why we work with a finite field instead of the reals.) Here the lines are used once more: at each horizontal coordinate the two possible heights are equally likely given the displayed point, and the heights at any two distinct coordinates are independent. This makes it unlikely that $h$ is accurate on an alternative column. The probability is small enough to sum over all alternative columns and all candidates disjoint from $F_m$. Thus, with high probability, the original column wins row $B$ for every such candidate, and all of them are discarded.

Now suppose that $D=F_m$. Every sheet in the hidden sequence is inside $F_m$. By Feature~\ref{feature}, the key of $F_m$ affects the line on none of these sheets, and none of their lists shows the tag of $F_m$. Changing the key of $F_m$ therefore changes neither the line nor the list on any hidden sheet, and every column gives exactly the same training examples. The scores can differ on other rows, but if we look at the table of scores without marking the original column, each column is still equally likely to be the original. Since each row selects just one column, the probability of discarding $F_m$ is at most $(N+1)/(100(N+2))<1/100$.

With high probability, then, $F_m$ survives and every surviving sheet shares at least one word with $F_m$. How do we find a word in $F_m$ without knowing which survivor it is? This is done by an additional structure we impose on the names of our sheets: First, all words in $U$ have the same length. Second, each sheet is the set of \emph{all} words that have particular letters at certain positions and may have any letter of the alphabet elsewhere. (Not every set of this form is a sheet, but every sheet has this form, as specified in Section~\ref{sec:family}.) We build our word one position at a time. If some surviving sheet requires a particular letter at that position, take the letter required by any such sheet. If none does, take any letter of the alphabet. To see why this works, consider a position where $F_m$ requires the letter $a$. The sheet $F_m$ itself supplies $a$, and any survivor requiring a letter there must also require $a$, since it shares a word with $F_m$. Thus the word we build satisfies every requirement for membership in $F_m$.

This completes our reduction from privately solving the word-finding task to our private PAC learning task, concluding the proof sketch of Theorem~\ref{main}. In Sections~\ref{sec:family}--\ref{sec:privacy}, we make this proof formal.

\paragraph{The proof of Theorem~\ref{thm:card}.}
The proof of Theorem~\ref{thm:card} builds on similar ideas. In particular, there is again a hidden nested sequence of $m$ sheets, hiding any one sheet of the sequence again leaves many equally likely candidates for it, and the lower bound again comes from the few examples that depend on that one sheet. However, the construction becomes somewhat more complex, as the class must now be small. The tags and lists used above are far too costly for this, and so is the family of sheets itself, whose size grows as a tower in $m$. Instead, a concept is the single word on the first hidden sheet, an input is a possible last sheet, and the label says whether it contains that word. The sheets come from a finite group, whose mixing properties replace the test table. The formal proof is given in Section~\ref{sec:card}.

\section{A family of sets of words}\label{sec:family}
The following coordinate construction is an iterated power-complex construction~\cite{DS}. Put $m=L-2$ and $r=(m+1)^4$, and write $[r]=\{1,\ldots,r\}$. Start from a singleton $X$ with family $\{\varnothing,X\}$. In each of $m$ steps replace $X$ by the set $[r]^X$ of words with positions indexed by $X$. If the old family is $\BB$, take as the new family the empty set and all sets of words of the form
\[
 C(S,a)=\{w\in[r]^X:w_i=a_i\text{ for every }i\notin S\},
 \qquad S\in\BB,\quad a\in[r]^X.
\]
We call these sets faces. Their free positions form an old set $S$. At the first step the family consists of the empty set, singletons, and the whole alphabet. At the second, words have $r$ positions and a proper nonempty face leaves at most one position free. Write $U,\BB$ for the final ground set and family.

\begin{lemma}\label{family}
The family $\BB$ is intersection-closed. All maximal chains from $\varnothing$ to a given member have the same number of strict inclusions, which we call the \emph{rank} of that member. In particular, every maximal chain of $\BB$ has $m+1$ strict inclusions. In an interval between two nested members whose ranks differ by two there are $r$ intermediate members, and any two intersect in the bottom of the interval. Of any three members, some two have the same join as all three. Here the join is the smallest member containing their union.
\end{lemma}
\begin{proof}
Two sets with conflicting prescribed letters are disjoint. If they contain a common word, their intersection keeps that word and intersects their free sets. This proves intersection closure by induction. Between nested nonempty sets the interval is exactly the old interval of free sets. Consequently the rank of a nonempty set is one more than the old rank of its free set, with the empty set at rank zero. A two-rank interval either comes from the previous step or starts at the empty set and ends at a set with one free position. The latter contains $r$ singletons. In either case, the intersection of distinct intermediate members must be the bottom by their ranks.

For three sets, some two have the same intersection as all three. If some pair is disjoint the assertion follows immediately. Otherwise their letter prescriptions are compatible, so all three contain a common word and the assertion follows by induction on their free sets. To see the join claim, suppose each of $B_1,B_2,B_3$ contributes to their join. Let $J_i$ be the join of the other two. An element of $B_i\setminus J_i$ lies in both other $J$'s. Thus removing any $J_i$ strictly enlarges the intersection of the three, a contradiction. The initial singleton family satisfies the induction claims.
\end{proof}

\section{A concept and a labeled example}
Index the parameters by \emph{all} nonempty proper members $\mathcal P=\BB\setminus\{\varnothing,U\}$. These are the $N$ sheets of Section~\ref{sec:idea}, so $|\mathcal P|=N$. Put $s=|\BB|=N+2$, $\kappa=100s$ and $k=\kappa^2$. Choose $q=2^d\ge sk$ and fix a set of $\kappa$ distinct keys in $\FF_q$. Give each nonempty $B\in\BB$ a public set $S_B$ of $k$ arguments in $\FF_q$, with these sets pairwise disjoint.

A target assigns to each $D\in\mathcal P$ a key $x_D$ and an affine function $A_D\colon\FF_q\to\FF_q$. We call the pair $(x_D,A_D)$ the \emph{tag} of $D$. The target also chooses one affine function $T$ shared by all bases. For $D\in\mathcal P$ and nonempty $B\in\BB$, put
\begin{equation}
 c_D=\sum_{E\in\mathcal P:\,E\cap D=\varnothing}x_E,
 \qquad f_B=T+\sum_{D\in\mathcal P:\,D\supseteq B}c_D A_D.
\end{equation}
The word $w_B$ lists the tags of the sets $D\in\mathcal P$ that do not contain $B$, in a fixed order. An input consists of $(B,w,a,b)$, where $\varnothing\ne B\in\BB$, $w$ has the format of $w_B$, $a\in S_B$, and $b\in\FF_q$. The target accepts exactly when $w=w_B$ and $b=f_B(a)$. Include one public dummy input labeled zero. These functions form $\HH_L$. At $B=U$, the word lists every tag and $f_U=T$. Since $S_U$ has at least two arguments, the inputs at $U$ distinguish all parameter assignments. In particular, $|\HH_L|=q^2(\kappa q^2)^{|\mathcal P|}$.

In the terminology of Section~\ref{sec:idea}, the members of $\mathcal P$ are the sheets and $U$ is the reference sheet. The tag of $D$ consists of its key $x_D$ and its adjustment $A_D$. The word $w_B$ is the list on sheet $B$, the graph of $f_B$ is its line, and $T=f_U$ is the line on the reference sheet. An argument $a\in S_B$ is a horizontal coordinate assigned to $B$, and $b$ is the height of the displayed point.

\paragraph{The distribution.} Choose a maximal chain $\varnothing=F_0\subsetneq F_1\subsetneq \cdots\subsetneq F_m\subsetneq F_{m+1}=U$ uniformly. Independently of the chain, draw all keys and affine functions independently and uniformly. These choices define the distribution. Each example is the labeled dummy with probability $1/2$. Otherwise choose $B$ uniformly from the $m$ internal chain members and $a$ uniformly from $S_B$. Write $z=f_B(a)$ in a fixed $d$-bit encoding, and let $b$ be $z$ with its last bit set to one. Give the input $(B,w_B,a,b)$ the original last bit of $z$ as its label. This is exactly the target label. The learner sees all but the last bit of the correct value and tries to predict that bit. The examples are independent given the chain and target.

\begin{lemma}\label{dimensions}
The class $\HH_L$ has VC dimension two and Littlestone dimension $m+2$.
\end{lemma}
\begin{proof}
For $\varnothing\ne B\subseteq C$, the difference $f_B-f_C$ is an affine function determined by $w_C$. Each term lost in passing to $C$ comes from a set $D$ visible at $C$, and every $E$ disjoint from $D$ misses the nonempty $B$ and is also visible at $C$. The words at $B,B'$ together determine the word at their join, since a set contains the join exactly when it contains both.

Suppose three distinct inputs are positive for some target. Two of their bases generate the full join $C$ by Lemma~\ref{family}. These two inputs fix $w_C$. They impose two evaluations on $f_C$ by the preceding difference identity. The arguments are distinct because different bases have disjoint argument sets, and two jointly positive inputs at the same base and argument must coincide. Hence $f_C$ is determined. The word and answer at the third base are now fixed, so every target accepting the pair accepts the third input. The class cannot shatter three inputs. For the lower bound, fix the keys and set every $A_D=0$. At one base use the correct word, choose two distinct arguments, and propose zero at both. The two values of $T$ can be specified independently, so all four labelings occur.

For the online upper bound, predict zero until the first positive input, then predict one only on that input until a second distinct positive arrives. As in the VC argument above, the two positives determine $w_C$ and $f_C$ at the join $C$ of their bases. The learner can now predict exactly at every base contained in $C$ and predict zero elsewhere, including on the dummy. Any further mistake is a positive at a base $B$ outside $C$. Put $C'=C\vee B$. The words $w_C$ and $w_B$ determine $w_{C'}$, which in turn determines $f_C-f_{C'}$, so the learner can update to $(C',w_{C'},f_{C'})$. Each such mistake strictly enlarges $C$. Since $C$ starts nonempty, Lemma~\ref{family} allows at most $m$ enlargements. This gives at most $m+2$ mistakes and hence $\LD(\HH_L)\le m+2$.

For the reverse bound use a full chain $R_1\subsetneq\cdots\subsetneq R_{m+1}=U$ starting at a singleton. Set every $A_D=0$, so $f_B=T$ everywhere. Fix all keys except $x_{R_1},\ldots,x_{R_m}$ to one public baseline value. At $R_1$ two different arguments with proposed answer zero can receive any of the four labelings by choosing $T$ on each branch. Their visible words contain only baseline keys. Then query $R_2,\ldots,R_{m+1}$ in order with the correct proposed answer $T(a)$. At $R_j$ the key $x_{R_{j-1}}$ becomes visible for the first time. Use its baseline value in the query word and the previously chosen values for earlier keys. Choosing the new key equal to that value or different gives either label. All later keys are still hidden. The resulting tree is shattered and has depth $m+2$.
\end{proof}

\paragraph{Why the ordinary cube does not suffice.} Consider using all subcubes of $\{0,1\}^m$ in place of the recursive family. For each $i$ take a query at the singleton $\{e_i\}$ with all visible keys at their baseline values, all visible functions zero, and proposed answer zero. Set $T$ and all $A_D$ to zero. The set $D_j=\{z:z_j=0\}$ contains all the unit vectors except $e_j$. Changing $x_{D_j}$ from its baseline value therefore flips only the $j$th label. These $m$ queries are shattered.

\section{Recovering a word from an arbitrary predictor}
For any nonempty $B\in\BB$ and any parameter assignment, define the \emph{accuracy at $B$} by drawing $a$ uniformly from $S_B$ and generating the missing-bit example at $B$. This is defined even for a base outside the sampled chain and can be computed by enumerating $S_B$. If the word and predictor are fixed and the correct function $f$ is a fresh uniform affine function, then
\begin{equation}\label{rare}
 \Pr\bigl[\text{accuracy at } B\ge 4/5\bigr]<3/k.
\end{equation}
Indeed, for each $a$, the last bit of the uniform value $f(a)$ is fair given its other bits. The correctness indicators at distinct arguments are pairwise independent because two values of a uniform affine function at distinct arguments are independent and uniform. Their average has mean $1/2$ and variance $1/(4k)$. Chebyshev's inequality at distance $3/10$ proves~\eqref{rare}.

\paragraph{The decoder.} The keys and functions are available to the decoder because we drew them independently of the chain. For each $D\in\mathcal P$, try all $\kappa$ values of $x_D$, keeping everything else fixed. At each nonempty base, guess the key maximizing the predictor's accuracy there, with fixed tie breaking. Discard $D$ if any base guesses its actual key.

\emph{The true set survives.} Fix the chain and let $I=F_m$. Its key $x_I$ occurs in no example. Indeed every sampled $B$ lies in $I$, so its word omits the tag of $I$. Every $D\supseteq B$ intersects $I$, so $c_D$ never reads $x_I$. Even given the other parameters and the predictor, $x_I$ remains uniform. The list of hypothetical accuracies is independent of which key is real. The probability that any base guesses it is at most $s/\kappa$.

\emph{Disjoint sets are discarded.} Condition on all keys, all the functions $A_D$ except $A_I$, and the single affine function
\[
                         M=T+c_I A_I.
\]
Because $T$ is independently uniform, $A_I$ is still uniform. For every sampled base the function $f_B$ contains $M$ as a summand, and every sampled word omits $A_I$. Thus conditioning further on the trained predictor leaves $A_I$ uniform. When total error is at most $1/10$, the average nondummy error is at most $1/5$. Choose the first chain base $B$ with accuracy at least $4/5$. This choice is determined by what we have conditioned on.

For any $D\in\mathcal P$ disjoint from $I$, change $x_D$ by $\Delta\ne0$ to a trial key. The resulting correct function at $B$ is
\begin{equation}
       f'_B=f_B+\Delta\!\sum_{\substack{J\in\mathcal P,\ J\supseteq B\\
                                      J\cap D=\varnothing}} A_J.
\end{equation}
The sum includes $A_I$ exactly once. All its other terms and the original $f_B$ are fixed. The trial word omits $I$, so it is fixed even though $A_I$ remains uniform. The new correct function is therefore uniform, so~\eqref{rare} applies to this trial. Across all disjoint $D$ and wrong keys there are at most $s\kappa$ trials. With probability at least $1-3s\kappa/k$, all these trials have accuracy below $4/5$ at this one good base. Consequently the decoder discards every $D$ disjoint from $I$.

The PAC guarantee gives total error at most $1/10$ with probability at least $9/10$. Thus with probability at least $9/10-s/\kappa-3s\kappa/k=0.86$, $I$ survives and every survivor intersects $I$. Fill each word position with any letter prescribed there by a survivor, using a default if none prescribes one. At every position fixed by $I$, its letter is available, and every survivor that prescribes a letter there agrees with $I$. The assembled word belongs to $I$. Fixed orderings give a definite output on every run.

\section{The privacy bound}\label{sec:privacy}
\begin{proof}[Proof of Theorem~\ref{main}]
By Lemma~\ref{dimensions}, the class $\HH_L$ has VC dimension two and Littlestone dimension $m+2=L$. It remains to prove the lower bound on $n$. We use replacement adjacency on databases of $n$ records. Suppose we receive only the underlying records, each either a dummy or a nonempty member of $\BB$. Draw the target parameters, convert each record separately to a labeled example, run the learner and apply the decoder. Conditional on the parameters and preprocessing coins, a replacement affects at most one training example. For each fixed choice of these independent coins and parameters, the learner is $(1,\delta)$-private and decoding is postprocessing. Averaging preserves privacy. On records drawn from the random chain, the resulting word lies in $F_m$ with probability at least $4/5$.

For $1\le j\le m$, replace the $N_j$ occurrences of level $j$ by dummies. Condition on which level or dummy was selected at each of the $n$ positions and on all chain members other than $F_j$. Then the erased procedure is independent of $F_j$. Each way of filling the gap completes exactly one maximal chain. Lemma~\ref{family} gives $r$ equally likely choices for $F_j$, and their differences from $F_{j-1}$ are disjoint. The erased output belongs to $F_j\setminus F_{j-1}$ with probability at most $1/r$. Restoring the records multiplies this bound by $e^{N_j}$ and adds $\delta(e^{N_j}-1)/(e-1)$ by group privacy (apply it first for the fixed full chain, then average over the missing member). The disjoint differences partition $F_m$. Since $N_j\sim\mathrm{Bin}(n,1/(2m))$, summing and taking expectations gives
\begin{equation}\label{eq:privacy}
 \frac45\le\frac{m\mu}{r}+\frac{m\delta}{e-1}(\mu-1),
 \qquad \mu=\left(1+\frac{e-1}{2m}\right)^n .
\end{equation}
When $\delta\le1/n$ and $n\le m/4$, we have $\mu<1.25$ and $\mu-1\le \mu(e-1)n/(2m)$. The right hand side is at most $1.25(m/r+1/2)<0.68$, since $m/r\le1/27$. Hence $n>m/4$. When $\delta=n^{-1.1}$ this gives $m\delta=O(m^{-0.1})$. The same inequality now requires $\mu=\Omega(m^{0.1})$, so $n=\Omega(m\log m)$. Since $m=L-2$, both claims follow. The linear bound also covers the finitely many small $L$ after adjusting the absolute constant in the superlinear bound.
\end{proof}

\paragraph{Dependence on $\delta$.} For a fixed privacy parameter $0\le\delta<1$, rearranging~\eqref{eq:privacy} and using $\log\mu\le(e-1)n/(2m)$ gives
\[
 n\ge\frac{2m}{e-1}\log\!\left(1+
      \frac{4/5-m/r}{m/r+m\delta/(e-1)}\right).
\]
Since $m/r=\Theta(L^{-3})$ and $4/5-m/r$ is bounded below by an absolute positive constant, this yields
\begin{equation}
 n=\Omega\!\left(L\log\!\left(1+\min\!\left\{L^3,\frac1{L\delta}\right\}\right)\right),
\end{equation}
with $1/(L\delta)=+\infty$ at $\delta=0$. In particular, the bound is $\Omega(1/\delta)$ for $\delta\ge1/L$ and $\Omega_\eta(L\log L)$ for $\delta\le L^{-1-\eta}$ and any fixed $\eta>0$.

\section{A lower bound in terms of class cardinality}\label{sec:card}
In this section we prove Theorem~\ref{thm:card}. The comparison of chambers that agree outside one rank (Section~\ref{sec:g-privacy}) is adapted from the erasure argument of Section~\ref{sec:privacy}.

\paragraph{Conventions.}
As in Section~\ref{sec:privacy}, we use replacement adjacency on databases of $n$ records, and we write $\mathcal A$ for the learner. As in the previous sections, $m$ is the length of the hidden sequence, $n$ is the number of samples, and $h$ denotes a predictor. The other symbols are specific to this section and to the appendices. In particular, $U$ now denotes the set of atoms defined below, which plays the role of the universe of words.

\subsection{The gallery class}\label{sec:g-class}
Set $\Lambda=6m$. The following rule generates a family of trees with vertex colors in $[m]$. Start with the singletons. Take a list of $2$ to $\Lambda$ previously generated trees, with disjoint vertex sets. Join consecutive trees by one edge each, provided that \emph{the color of each endpoint is absent from the opposite tree in that pair}. Keep the resulting tree. We call the trees so generated \emph{tests}.

Every connected subtree of a test is a test. Indeed, it meets a consecutive interval of the pieces in connected subtrees, and the joining edges still satisfy the rule. Induction on the construction proves the assertion. A test using $s$ colors has at most $\Lambda^{s-1}$ vertices. In a nontrivial join each piece omits an opposite endpoint color, so it uses at most $s-1$ colors, and there are at most $\Lambda$ pieces. In particular the tests are finite in number. Put
\[
 C=\Lambda^{m-1},\quad W=(4m)^{C+1},\quad p_0=(100m)^{10m\,2^C},\quad p_0<p<2p_0,
\]
where $p$ is prime, chosen using Bertrand's postulate.

Choose one root per test type. For a word $w$ over $[m]$, let $w(Q)$ count modulo $p$ the placements of a rooted test $Q$ in $w$ that match colors and put each parent strictly before its children. Vertices on separate branches may occupy the same position. Identify words if these counts agree on every chosen test. Appendix~\ref{sec:g-group} shows that the chosen roots determine all rootings and that concatenation gives a finite $p$-group $G$ of nilpotence class at most $C$, with $|G|\le p^W$. Write $a_i$ for the one-letter word of color $i$, $H_i=\langle a_i\rangle$ for its order-$p$ subgroup, and $\pi_i$ for the homomorphism deleting color $i$.

Let $P_i=\pi_i(G)$ be the subgroup generated by all colors except $i$. An $i$-vertex is a left coset $gP_i$. Vertices of different types are adjacent when their cosets intersect. A chamber is the tuple $(gP_1,\ldots,gP_m)$ for $g\in G$. Orient each edge from higher to lower type. A descending gallery follows adjacent vertices of consecutive decreasing types. Directed paths have the same reachability as galleries, since a representative of the intersection on any edge supplies all skipped types in its chamber. Appendix~\ref{sec:g-group} excludes induced cycles of lengths $4$ through $6m$. For a vertex $v$ let $F(v)$ be the set of type-one vertices reachable this way, allowing a stationary gallery. These type-one vertices form the atom set $U=G/P_1$. The inputs are the top vertices $G/P_m$ and the finite binary class is
\[
 \HH'_m=\{h_x:x\in U\},\qquad h_x(v)=\mathbf1\{x\in F(v)\},
\]
with identical functions counted only once.

\paragraph{A path test and the density of incidence.}
A gallery from $gP_m$ to $g'P_1$ gives $g^{-1}g'\in P_mP_{m-1}\cdots P_1$, by taking a representative at each intersection. No word formed from successive factors omitting $m,m-1,\ldots,1$ can contain $m,m-1,\ldots,1$ as a subsequence. The first marked letter must occur after the first factor, the second after the second factor, and so on. The decreasing path itself is a test, formed by joining singletons. Thus a nonzero count for a group element $u$ on this path rules out $u\in P_m\cdots P_1$.

Write $\phi(u)$ for the decreasing-path count of $u\in G$. Fix $u$ and append the word $w_t=a_m^t a_{m-1}^t\cdots a_1^t$ for $0\le t<p$. Then $\phi(uw_t)$ is a monic polynomial of degree $m$ in $t$. Placements using exactly $j$ letters from $w_t$ contribute $t^j$ times the number of placements of the first $m-j$ colors in $u$, with leading coefficient one from $j=m$. Thus $\phi(uw_t)=0$ for at most $m$ values of $t$. For independent uniform $u$ and $t$, $uw_t$ is uniform in $G$ by right translation, so $\Pr[\phi(u)=0]\le m/p$. With either the atom or the top fixed, a uniform representative for the other makes the relative element uniform in $G$. Hence the incidence probability is at most $m/p$ in both cases.

\paragraph{VC dimension and cardinality.}
Put $v_s=a_m^sP_m$ and $x_t=a_m^t a_{m-1}\cdots a_1 P_1$ for $s,t\in\{0,\ldots,p-1\}$. Then $v_s$ reaches $x_t$ exactly when $s=t$. For $s=t$ the cosets intersect. Otherwise their relative element has decreasing-path count $t-s\ne0$ in $\FF_p$. The $p$ atoms therefore give distinct concepts. The atom $P_1$ is reached by every $v_s$ because $a_m\in P_1$. On $v_0,v_1$ the atoms $P_1,x_0,x_1,x_2$ give all four labelings, so the VC dimension is at least two, and
\begin{equation}\label{eq:g-size}
                       p\le|\HH'_m|\le|G|\le p^W .
\end{equation}

For the upper bound, a shattered triple would give tops $v_i$ and atoms $x_j$ with $v_i\leadsto x_j$ exactly for $i\ne j$. The graph induced by six witnessing galleries has at most $6m$ vertices and is chordal. Label each of its vertices $z$ by $\operatorname{lab}(z)=\{i:v_i\leadsto z\}$ using reflexive reachability. The label is nonempty, and it omits $j$ whenever $z$ lies on a witnessing gallery ending at $x_j$. Labels increase by inclusion along a directed edge. The six nonempty proper subsets of $\{1,2,3\}$, with distinct comparable sets joined, form a hexagon. A triangle in the witness graph uses at most two labels, so its image crosses each hexagon edge an even number of times. Every closed walk has the same parity property, by splitting at repeated vertices and chords. Yet the six galleries concatenate in alternating directions into a closed walk that crosses each hexagon edge exactly once. Indeed, a gallery from $v_i$ to $x_j$ changes its label just once, from $\{i\}$ to $\{1,2,3\}\setminus\{j\}$. This contradiction proves the upper bound. Hence $\VC(\HH'_m)=2$.

\subsection{Accurate predictors and cosets}\label{sec:g-cov}
We use one estimate for real functions on a coset. Let subgroups $X,Y$ generate $K\le G$, and draw a uniform point of a left coset $uK$. If $f$ is constant on the $X$-cosets and $f'$ on the $Y$-cosets in $uK$, then
\begin{equation}\label{eq:g-cov}
 |\operatorname{Cov}(f,f')|\le\xi\sqrt{\operatorname{Var}(f)\operatorname{Var}(f')},
 \qquad \xi=2p^{-1/2^{C-1}}.
\end{equation}
To obtain~\eqref{eq:g-cov}, apply Lemma~\ref{lem:g-angle} of Appendix~\ref{sec:g-angle} to right translation by $K$ on the mean-zero functions in $L^2(uK)$ with uniform measure. Only constants are fixed by $K$, since this action on $uK$ is transitive. The centered functions $f-\mathbb Ef$ and $f'-\mathbb Ef'$ are fixed by $X$ and $Y$, respectively. Their inner product is the covariance, their squared norms are the variances, and $K$ has nilpotence class at most $C$.

\paragraph{A realizable distribution for each chamber.}
Choose $g$ uniformly in $G$. For the population $\mathsf P_g$, draw a top vertex $v$ in one of two ways. With probability $1/2$ draw $v$ uniformly from all tops. Otherwise choose $i$ uniformly from $[m-1]$, choose $y$ uniformly in $P_i$, and set $v=gyP_m$. In both branches the label is $h_{gP_1}(v)$. In the second branch this label is always one because the chamber of $gy$ descends to $gP_i$, and the chamber of $g$ descends from there to the atom $gP_1$. Conditional on $g$, the training examples are independent draws from the realizable distribution $\mathsf P_g$.

\paragraph{Where a fixed predictor is accurate.}
Write $f(g)=h(gP_m)\in[0,1]$ for a predictor's probability of predicting one and $b$ for its uniform mean. Write $f_i(g)$ for the mean of $f$ on $gP_i$. The mean in the second branch is $\bar b(g)=(m-1)^{-1}\sum_{i<m} f_i(g)$. The error differs from $\tfrac12(1-\bar b(g)+b)$ by at most $m/(2p)$. Indeed, assigning label zero to every uniform top changes an error only on the true atom's positive region. Hence error at most $0.10$ implies $\tfrac12(1-\bar b(g)+b)\le0.11$. Define
\[
                       A_h=\{g:\bar b(g)-b\ge0.78\}.
\]
For every fixed $g$, the PAC guarantee gives $\Pr_{h\sim\mathcal A(\mathsf P_g^n)}[g\in A_h]\ge0.90$. Since $f_i$ is the conditional mean of $f$ given $gP_i$, under uniform $g$ one has $\operatorname{Cov}(f_i,f)=\operatorname{Var}(f_i)$. The pair $P_i,P_m$ generates $G$, and $f$ is constant on $P_m$-cosets. Applying~\eqref{eq:g-cov} gives $\|f_i-b\|_2\le\xi\|f-b\|_2\le\xi$. Averaging over $i$ and applying Chebyshev gives
\[
                |A_h|/|G|\le \|\bar b-b\|_2^2/0.78^2\le 2\xi^2 .
\]

\paragraph{Assigning a rank.}
Set $\theta=m^{-4}$. Call each coset $gH_i$ a rank-$i$ panel. We show that a nonempty set $\Gamma\subseteq G$ occupying at least a $\theta$ fraction of each coset $gH_i$ it meets has density at least $\theta^m-2\xi$. Let $G_j=\langle H_1,\ldots,H_j\rangle$ and let $\gamma_j$ be the least fraction occupied by $\Gamma$ in any $G_j$-coset it meets. Thus $\gamma_1\ge\theta$ and $\gamma_m=|\Gamma|/|G|$. For $j\ge2$ take an occupied $G_j$-coset of density $\gamma=\gamma_j$. Within it, let $\Gamma',\Gamma''$ be the unions of the occupied $G_{j-1}$-cosets and the occupied $H_j$-cosets, respectively. Their relative sizes $\gamma',\gamma''$ satisfy $\gamma'\le \gamma/\gamma_{j-1}$ and $\gamma''\le \gamma/\theta$. Every point of $\Gamma$ there belongs to $\Gamma'\cap \Gamma''$. Applying~\eqref{eq:g-cov} to the indicators gives
\[
  \gamma\le \gamma'\gamma''+\xi\sqrt{\gamma'\gamma''}
  \le\frac{\gamma^2}{\theta \gamma_{j-1}}+
             \frac{\xi \gamma}{\sqrt{\theta \gamma_{j-1}}}.
\]
Dividing by $\gamma>0$ and rearranging yields $\gamma_j\ge\theta \gamma_{j-1}-\xi\sqrt{\theta \gamma_{j-1}}\ge\theta \gamma_{j-1}-\xi$. Iterating and using $\theta\le1/2$ gives the claimed density. Here $\xi\le2(100m)^{-20m}$. Thus $\theta^m-2\xi>2\xi^2$ for large $m$. Hence $A_h$ contains no such $\Gamma$.

In particular, every nonempty subset of $A_h$ meets some panel $gH_j$ in at least one and fewer than $\theta p$ chambers. Repeatedly remove one such chamber and assign it that panel's rank, with fixed public orders for ties. At first use a panel already has fewer than $\theta p$ chambers remaining. This defines a map $j_h:A_h\to[m]$ with
\begin{equation}\label{eq:g-charge}
     |\{g\in uH_j\cap A_h:j_h(g)=j\}|<\theta p\qquad(u\in G,\ j\in[m]).
\end{equation}

\subsection{Privacy and parameter bounds}\label{sec:g-privacy}
\paragraph{Privacy for product distributions.}
Let $\mathsf P,\mathsf Q$ be distributions on the same finite example space with $\operatorname{TV}(\mathsf P,\mathsf Q)\le \tau\le 1$. For any replacement $(1,\delta)$-private algorithm $\mathcal A$ on $n$ records and any output event $E$,
\[
 \Pr[\mathcal A(\mathsf P^n)\in E]\le \mu\Pr[\mathcal A(\mathsf Q^n)\in E]+\delta\,(\mu-1)/(e-1),
 \qquad \mu=(1+(e-1)\tau)^n .
\]
\begin{proof}
Fix all but one record, and let $\Phi(x)$ be the probability of $E$ when the remaining record is $x$. By replacement privacy, $\max \Phi-\min \Phi\le(e-1)\min \Phi+\delta$. Hence
\[
 \mathbb E_{\mathsf P} \Phi-\mathbb E_{\mathsf Q} \Phi
 \le \tau(\max \Phi-\min \Phi)\le \tau\big((e-1)\mathbb E_{\mathsf Q} \Phi+\delta\big).
\]
Changing one marginal from $\mathsf Q$ to $\mathsf P$ thus takes an event probability $t$ to at most $(1+(e-1)\tau)t+\tau\delta$, also after averaging over the other independent records. Change the $n$ marginals in turn and sum the geometric progression.
\end{proof}

\paragraph{Comparing chambers in one panel.}
Let $g'\in gH_j$. On the second branch of $\mathsf P_g$ every rank $i\ne j$ has the same law as under $\mathsf P_{g'}$, since $H_j\subseteq P_i$. Only the branch at rank $j<m$ can change. On the uniform-top branch, couple the same top in both laws. The labels differ only in the union of the two true atoms' positive regions, of mass at most $2m/p$. It follows that
\[
 \operatorname{TV}(\mathsf P_g,\mathsf P_{g'})\le \tau:=\frac{1}{2(m-1)}+m/p,
 \qquad \mu=(1+(e-1)\tau)^n.
\]
Draw $g$ and $J$ independently and uniformly from $G$ and $[m]$, then draw $g'$ uniformly from $gH_J$. Apply the product-distribution bound to $\mathsf P_g,\mathsf P_{g'}$ with the event $E_{g,J}=\{h:g\in A_h,\ j_h(g)=J\}$, keeping $g,g',J$ fixed for this comparison. Average over these choices. Under $\mathsf P_g^n$ the event has averaged probability at least $0.90/m$, by the PAC guarantee and the independent choice of $J$, since each $g\in A_h$ is assigned exactly one rank. Under $\mathsf P_{g'}^n$, conditional on $g',J$, the original $g$ is uniform in $g'H_J$ independently of the learner's output. The charge bound~\eqref{eq:g-charge} gives averaged probability at most $\theta$. Therefore
\begin{equation}\label{eq:g-privacy}
             0.90\le m\mu \theta + m\delta(\mu-1)/(e-1).
\end{equation}
If $\delta\le1/n$ and $n\le m/8$, then $m\tau\to1/2$ implies $\mu\le\exp((e-1)n\tau)<1.2$ for large $m$. Also $\mu-1\le\mu(e-1)n\tau$. The right-hand side of~\eqref{eq:g-privacy} is then at most $1.2(m^{-3}+m\tau)<0.8$ for large $m$. This proves $n>m/8$. For $\delta\le n^{-1.1}$, the linear bound gives $m\delta=O(m^{-0.1})$. As in the proof of Theorem~\ref{main}, inequality~\eqref{eq:g-privacy} now forces $\mu=\Omega(m^{0.1})$, so $n=\Omega(m\log m)$.

Finally, $\log\log p=\Theta(C)$ and $\log W=O(C\log m)$. By~\eqref{eq:g-size}, $\log\log|\HH'_m|$ lies between $\log\log p$ and $\log\log p+\log W$. Taking one more logarithm gives $\log\log\log|\HH'_m|=\Theta(\log C)=\Theta(m\log m)$. This completes the proof of Theorem~\ref{thm:card}.

\section*{AI disclosure}
The initial constructions and proofs in this work were obtained through extended interactions with ChatGPT, Claude, and Gemini. Throughout this process, the authors maintained a research roadmap that tracked which directions were still under consideration and which had been ruled out. The authors then worked to simplify the constructions and proofs. The authors take full responsibility for the paper's content.

\appendix
\section{The group defined by rooted tree counts}\label{sec:g-group}
This appendix and the next use the notation of Section~\ref{sec:card}. For adjacent vertices $\rho,\rho'$ of a test $Q$, write $Q_\rho,Q_{\rho'}$ for the components obtained by cutting their edge. Both are tests. The colors of $\rho,\rho'$ differ, so their two possible orders give
\begin{equation}\label{eq:g-reroot}
  w(Q,\rho)+w(Q,\rho')=w(Q_\rho,\rho)\,w(Q_{\rho'},\rho'),
\end{equation}
where the second argument indicates the root. Induction on size recovers every rooting from the chosen ones. Extend counts to forests by multiplying their component counts, with value one on the empty forest. This counts placements because distinct components can use positions independently.

\paragraph{Concatenation and the group law.}
For words $u,u'$ and a forest $\Psi$,
\begin{equation}\label{eq:g-concat}
 (uu')(\Psi)=\sum_{I\sqcup J=V(\Psi)}u(\Psi[I])\,u'(\Psi[J]),
\end{equation}
where the sum is over ordered partitions such that each edge between the parts points from $I$ to $J$. Indeed $I$ is exactly the set of vertices placed in the prefix $u$. Orient edges away from the roots for this formula. Each induced component inherits a root at its vertex nearest the original root and remains a test. Thus the tree counts of $uu'$ are determined by those of $u$ and $u'$, and concatenation is well defined on equivalence classes. The word consisting of $p$ copies of a single color $i$ has the same counts as the empty word. On singletons the count is zero modulo $p$, and a tree with at least one edge needs at least two colors. It follows that every one-letter word has an inverse and has order exactly $p$ by its singleton count. All word classes therefore form a group.

There are at most $W=(4m)^{C+1}$ chosen rooted tree types. Indeed, on $j\le C$ vertices, ordering the children of a rooted tree gives at most $4^{j-1}$ tree shapes, by the parentheses encoding of a depth-first traversal. Allow $m^j$ colorings and sum over $j$. Consequently $|G|\le p^W$. To see also that $G$ is a $p$-group and to bound its nilpotence class, let $G^{(d)}$ be the kernel of the counts on trees of size less than $d$, meaning the word classes that agree with the identity on these trees. Truncating the counts gives another group quotient by \eqref{eq:g-concat}, so each $G^{(d)}$ is normal. For $g\in G^{(d)}$ and a tree $Q$ of size $d$, the mixed terms of \eqref{eq:g-concat} vanish, since the vertices assigned to $g$ would give a nonempty forest of size less than $d$. Thus $(gg')(Q)=g(Q)+g'(Q)=(g'g)(Q)$ for every $g'\in G$, with the same equality at smaller sizes. The degree-$d$ count vector embeds $G^{(d)}/G^{(d+1)}$ into an additive group over $\FF_p$, and this quotient is central in $G/G^{(d+1)}$. Since $G^{(1)}=G$ and $G^{(C+1)}=\{1\}$, the assertion follows.

Deleting color $j$ preserves all counts on trees missing $j$ and sets the other tree counts to zero. Thus deletion induces the retraction $\pi_j:G\to P_j$. Deletions commute.

\paragraph{Detecting a product of two subgroups.}
For colors $j,k$, possibly equal, $g\notin P_jP_k$ exactly when some rooted test with root color $j$ and a vertex of color $k$ has nonzero $g$-count. Indeed, in a word missing $j$ followed by a word missing $k$, the root would have to lie in the second part, forcing every vertex there.

Conversely, suppose $g\notin P_jP_k$. The element
\[
    \hat g=\pi_j(g)\,[\pi_j\pi_k(g)]^{-1}\,\pi_k(g)\in P_jP_k
\]
agrees with $g$ after deletion of either color. Choose a smallest tree on which $g,\hat g$ differ. It must use both colors. Reroot at a $j$-vertex. Each change across an edge negates the count difference by~\eqref{eq:g-reroot}, since $g,\hat g$ agree on smaller trees. The rerooted tree has $\hat g$-count zero and $g$-count nonzero, as required.

\paragraph{Excluding short cycles.}
Suppose vertices $z_1,\ldots,z_n$ form a chordless cycle of cosets of types $s_1,\ldots,s_n$, with $4\le n\le \Lambda$. With cyclic indices choose $y_i\in z_i\cap z_{i+1}$ and put $g_i=y_{i-1}^{-1}y_i\in P_{s_i}$. Their product is the identity. If $g_i=\sigma\sigma'$ with $\sigma\in P_{s_{i-1}}$, $\sigma'\in P_{s_{i+1}}$, the point $y_{i-1}\sigma=y_i\sigma'^{-1}$ belongs to both neighboring cosets. This would give a chord if their types differ and identify two distinct cycle vertices if their types agree. Choose a smallest test $Q_i$ of nonzero $g_i$-count whose root $\rho_i$ has color $s_{i-1}$ and which contains a vertex $\beta_i$ of color $s_{i+1}$. Such a test exists by the preceding criterion. It omits $s_i$ because $g_i\in P_{s_i}$. Write $d_i=|V(Q_i)|$.

For $i<n$ attach $\rho_{i+1}$ as a child of $\beta_i$. The result $Q$, rooted at $\rho_1$, is a test by the construction rule. The color $s_{i+1}$ of $\beta_i$ is absent from $Q_{i+1}$, and the color $s_i$ of $\rho_{i+1}$ is absent from $Q_i$.

By~\eqref{eq:g-concat}, in each nonzero term of the count of $g_1\cdots g_n$ on $Q$, factor indices are nondecreasing along the spine $\rho_1,\beta_1,\ldots,\rho_n,\beta_n$. The index of $\rho_i$ is at least $i$ by induction, since $\rho_{i+1}$ has color $s_i$, absent from factor $i$. Conversely the index of $\beta_i$ is at most $i$ by backward induction, since $\beta_i$ has color $s_{i+1}$, absent from factor $i+1$. Thus the entire path $\rho_i\to \beta_i$ lies in factor $i$. The component of that factor containing the path is rooted at $\rho_i$, since the parent of $\rho_i$, when present, belongs to factor $i-1$. Its count is nonzero, since the forest count is the product over components. It contains the marked colors in the required positions, so minimality forces it to have at least $d_i$ vertices. As $Q$ has exactly $\sum_i d_i$ vertices, factor $i$ has exactly $d_i$. Every vertex of $Q_i$ descends from $\rho_i$, hence lies in factor $i$ or later. Successively, factor $1$ must receive $Q_1$, factor $2$ must receive $Q_2$, and so on. The unique nonzero term contributes $\prod_i g_i(Q_i)\ne0$, contradicting the identity product. This proves the cycle exclusion. Related coset-cycle conditions are studied by Otto~\cite{Otto}.

\section{The subgroup-angle bound}\label{sec:g-angle}
We give the finite $p$-group case of the subgroup-angle bound of Caprace and Kassabov~\cite[Proposition~3.1]{CK}, with the constant explicit. The proof uses the rank-one projections of~\cite[Theorem~10.8]{EJK}.

\begin{lemma}\label{lem:g-angle}
Let $K=\langle X,Y\rangle$ be a finite $p$-group of nilpotence class $c\ge1$. In a unitary representation of $K$ with no nonzero $K$-fixed vectors, let $v,w$ be unit vectors fixed by $X,Y$, respectively. Then $|\langle v,w\rangle|\le \zeta_c$, where $\zeta_c=2p^{-1/2^{c-1}}$. If $K$ is abelian, $\langle v,w\rangle=0$.
\end{lemma}
\begin{proof}
By orthogonal decomposition and Cauchy--Schwarz it suffices to treat each nontrivial irreducible representation. For $c=1$ these are characters, and none is trivial on both $X$ and $Y$. For $c\ge2$, let $Z$ be the last nontrivial term of the lower central series. It is central and lies in $[K,K]$. If $Z$ acts trivially, apply induction to $K/Z$, of class $c-1$. Otherwise some element of $Z$ acts by a nonidentity scalar $\lambda$ by Schur's lemma. Writing $d$ for the dimension, $\lambda^d=1$ because the determinant is trivial on $[K,K]$. The order of $\lambda$ is a positive power of $p$, so $d\ge p$.

Conjugation on traceless matrices factors through $K/Z$ and has no nonzero fixed vector by Schur's lemma. Put $M_v=vv^*-I/d$ and $M_w=ww^*-I/d$. These matrices are fixed by $X,Y$, respectively. In the Hilbert--Schmidt inner product, $\|M_v\|^2=\|M_w\|^2=1-1/d$ and $\langle M_v,M_w\rangle=|\langle v,w\rangle|^2-1/d$. Induction gives
\[
 |\langle v,w\rangle|^2\le d^{-1}+(1-d^{-1})\zeta_{c-1}
 \le p^{-1}+\zeta_{c-1}\le \zeta_c^2. \qedhere
\]
\end{proof}
\end{document}